\documentclass[conference]{IEEEtran}
\IEEEoverridecommandlockouts

\usepackage{cite}
\usepackage{amsmath,amssymb,amsfonts}
\usepackage{algorithmic}
\usepackage{graphicx}
\usepackage{textcomp}
\usepackage{xcolor}
\usepackage{hyperref}
\usepackage{multirow}

\newcommand{\linebreakand}{
    \end{@IEEEauthorhalign}
    \hfill\mbox{}\par
    \mbox{}\hfill\begin{@IEEEauthorhalign}
}

\begin{document}
\begin{sloppypar}

\title{Deep Multi-View Learning for Tire Recommendation}

\author{
    \IEEEauthorblockN{Thomas Ranvier \href{https://orcid.org/0000-0001-9250-9530}{\includegraphics[scale=.2]{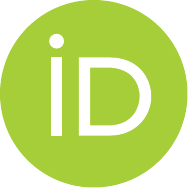}}}
    \IEEEauthorblockA{\textit{Université Lyon1 - LIRIS} \\
    43 bd du 11 Novembre 1918 \\
    69622 Villeurbanne, France \\
    \href{mailto:ranvier.thomas.pro@gmail.com}{ranvier.thomas.pro@gmail.com}}
    
    \and
    \IEEEauthorblockN{Khalid Benabdeslem}
    \IEEEauthorblockA{\textit{Université Lyon1 - LIRIS} \\
    43 bd du 11 Novembre 1918 \\
    69622 Villeurbanne, France \\
    \href{mailto:khalid.benabdeslem@univ-lyon1.fr}{khalid.benabdeslem@univ-lyon1.fr}}
    \\
    \IEEEauthorblockN{Bruno Canitia}
    \IEEEauthorblockA{\textit{Lizeo IT} \\
    42 Quai Rambaud \\
    69002 Lyon, France \\
    \href{mailto:bruno.canitia@lizeo-group.com}{bruno.canitia@lizeo-group.com}}
    
    \and
    \IEEEauthorblockN{Kilian Bourhis}
    \IEEEauthorblockA{\textit{Lizeo IT} \\
    42 Quai Rambaud \\
    69002 Lyon, France \\
    \href{mailto:kilian.bourhis@lizeo-group.com}{kilian.bourhis@lizeo-group.com}}
}

\maketitle

\begin{abstract}
We are constantly using recommender systems, often without even noticing.
They build a profile of our person in order to recommend the content we will most likely be interested in.
The data representing the users, their interactions with the system or the products may come from different sources and be of a various nature.
Our goal is to use a multi-view learning approach to improve our recommender system and improve its capacity to manage multi-view data.
We propose a comparative study between several state-of-the-art multi-view models applied to our industrial data.
Our study demonstrates the relevance of using multi-view learning within recommender systems.
\end{abstract}

\begin{IEEEkeywords}
multi-view, recommender system, attention mechanisms, deep learning
\end{IEEEkeywords}

\section{Introduction}

Many datasets contain data of various natures or coming from different sources, those are called multi-view data.
Multiple views can provide additional information compared to a single view, intelligently exploiting the multi-view aspect of such data can lead to an improvement in learning performance.
The main challenge of multi-view learning is not only to exploit the consensus knowledge contained in all views but also to exploit the complementarity of the views to improve performance.

In this paper we present a comparative study in a real-life application context on industrial data between a Baseline model that does not exploit the multi-view aspect of the data and three state-of-the-art models taking advantage of the multi-view aspect of the data.
We provide several results of different models showing the advantage of multi-view learning.
We thus demonstrate the relevance and interest of multi-view learning on a tire recommendation problem in an industrial context.

\subsection{Recommender systems}

The purpose of a recommender system is to estimate the products that are most likely to be of interest to a user \cite{betru_deep_2017}.
There exists two main recommendation tasks: the prediction of a rating and the creation of a ranking. 
The prediction of a rating is aimed at estimating the score that the user could give to a product and is generally based on the scores given to other products by the same user.
The point of creating a ranking is to define a ranked list composed of $n$ products ordered by the user's estimated taste preference.
It is usually called a ``top-$n$'' \cite{zhang_deep_2019}.
There are three main recommender system categories \cite{adomavicius_toward_2005}:
\begin{enumerate}
    \item \text{Content-based recommendations.} The user is recommended products similar to those that he has previously consulted and appreciated \cite{betru_deep_2017}.
    Recommendations are built from the user profile, they are based on the similarity between the products the user is interested in.
    To decide which product to recommend the system calculates a utility value $util(u, v)$ between the user $u$ and an item $v$ \cite{adomavicius_toward_2005}, usually defined as follows:
    \begin{equation}
        util(u,v) = score(profile(u), content(v))
    \end{equation}
    The system creates a profile for each user containing its tastes and preferences, noted $profile(u)$.
    It does the same for each product, noted $content(v)$.
    The two main drawbacks of this type of recommender system are the cold start and overspecialization problems.
    Indeed, it is hard to make pertinent recommendations to an unknown user, since its profile is empty, and conversely it is hard to properly recommend a new item that nobody has ever consulted or purchased, this is called the cold-start problem \cite{schein_methods_2002}.
    Overspecialization can also appear, this has the effect of enclosing the recommendations made to the user in a bubble and does not allow the recommendation of new and original products.
    \item \text{Collaborative Filtering.} The user is recommended products that people with similar taste have liked before.
    In that case the utility value $util(u, v)$ between the user $u$ and the item $v$ is estimated based on the utility values $util(u_j,v)$ assigned $v$ by users $u_j \in U$ which are similar to $u$ \cite{adomavicius_toward_2005}.
    In this way, users with similar profiles will be recommended similar products.
    That type of recommender system faces the cold-start problem too but is less subject to the overspecialization problem than content-based systems.
    It also faces the sparsity problem, since it is based on the rating of products by users and that each user only rates a very small part of available products.
    \item \text{Hybrid approaches.} The point of hybrid approaches is to combine the two previously described approaches together.
    The main point of combining content-based recommendations and collaborative filtering is to avoid specific limitations presented by each of those approaches \cite{balabanovic_fab_1997, basu_recommendation_1998}.
    The most common approach in recent years has been to design a model combining the two approaches \cite{adomavicius_toward_2005}, thus making it possible to solve, at least partially, the cold-start and sparsity problems \cite{betru_deep_2017}.
\end{enumerate}
A lesser known and used category is \textit{session-based} recommender systems \cite{tuan_3d_2017}.
Those use the similarities between user's browsing habits and their relationship to the consulted products to produce a recommendation.
In that case it is important to take into account the sequential nature of browsing sessions data.
It is therefore needed to use a model that is able to handle a temporal dimension, such as an RNN (Recurrent Neural Network) \cite{hidasi_parallel_2016} or a 3D-CNN (3D Convolutional Neural Network) \cite{tuan_3d_2017}.

\subsection{Multi-view learning}

A lot of data is collected from distinct sources.
Making intelligent use of these different views can improve the performance of a classification system.
Multi-view learning can also be applied to improve performance on data organized in a single view by manually generating multiple views \cite{zhao_multi-view_2017}.

In order for multi-view learning to be effective it is necessary that the views adhere to certain principles, otherwise the use of multiple views could lead to performance degradation.
The two main principles of multi-view learning are the consensus principle and the complementary principle \cite{xu_survey_2013}:

\begin{enumerate}
    \item \text{Consensus principle.}
    In supervised learning, this principle defines that by minimizing disagreements between each view, the error rate on each view will also be minimized.
    In other words, a system capable of correlating the results obtained from different views will at the same time minimize classification errors on each of these views.
    \item \text{Complementary principle.}
    This principle defines that in the context of multiple views, each view of the data contains knowledge that the others do not possess, thus, multiple views can be exploited to comprehensively and accurately describe the data and lead to better learning performance.
\end{enumerate}

By exploiting these two principles, it is theoretically possible to ensure a learning performance improvement by comparison to single view learning.

\subsection{Multi-view recommender systems}

A recommender system can benefit from multi-view learning provided that the used data allows the application of the consensus and complementary principles.
Concrete applications of the multi-view approach in the field of recommendation usually make use of deep learning composite models.
Several state-of-the-art models have been considered to be applied to our industrial data and be part of our comparative study.

\begin{itemize}
    \item Wide \& Deep model \cite{cheng_wide_2016}:
    This model combines a linear model with a deep model, allowing the benefits of each model to be exploited within a unified system.
    Deep models are capable of generalizing to unknown data but tend to overgeneralise when insufficient training data is used.
    Linear models are capable of storing exception rules on scattered data with a lower generalization capacity but a lower need for the number of parameters to be learned.
    The wide component of the model applies the linear model to a set of features, which can be raw or transformed with simple operations, such as vector products, and the deep component is a deep forward propagation neural network.
    Before being passed to the deep network, the scattered, high-dimensional data is converted into smaller, dense vectors, called embedding vectors.
    The model is trained as a whole using joint training, which allows multi-branch models to be trained using a single error function \cite{webb_joint_2019}.
    \item DeepFM model \cite{guo_deepfm_2017}:
    This model is based on the observation that the Wide \& Deep model requires extensive manual analysis work on the data in order to determine the best mathematical transformations to use for the wide component.
    The DeepFM model automates this manual work using a factorization machine (FM), which allows the modeling of pairwise interactions between data features using vector products between embedding vectors.
    The deep component of the model remains unchanged compared to that of the Wide \& Deep model.
    The model is also driven as a single model using joint training.
    \item NeuMF model \cite{he_neural_2017}:
    This model generates two latent vectors from the user's view and two others from a product view.
    The two vectors from each view are generated with matrix factorization (MF vector) and with a multi-layer perceptron (MLP vector).
    The NeuMF model shares similarities with the DeepFM, both models were published less than five months apart.
    \item MV-DNN model \cite{elkahky_multi-view_2015}:
    The core of the MV-DNN is a deep neural network (DNN) which is applied to several views.
    Its structure is essentially the same as that of a Deep Semantic Similarity Model (DSSM) \cite{gao_modeling_2014}, the difference is that the number of branches of the model can be adapted to the number of views of the data rather than being limited to two.
    Each view passes through a different DNN and the similarity between the outputs of each DNN is measured using a cosine similarity function.
    DSSMs are usually used on textual data but can be adapted to other types of data.
    Their architecture can be linear as in \cite{elkahky_multi-view_2015} or use another type, such as a convolutional structure in \cite{gao_modeling_2014}.
    \item TDSSM \cite{song_multi-rate_2016}:
    This model introduces a temporal dimension to the DSSM, it is called Temporal DSSM (TDSSM).
    The temporal dimension is added to the structure with a recurrent neural network (RNN), which makes the model capable of handling temporal views in addition to static views.
    In its application on a recommender system the user and product characteristics are passed as different views and the outputs of the neural networks are combined and compared with a cosine similarity function.
    This underlines the fact that the structure of a DSSM is fully modular and can be modified at will to meet data requirements.
    \item MV-AFM model \cite{liang_multi-view_2020}:
    The Multi-View Attentional Factorization Machines (MV-AFM) model is a neural network that uses hierarchical attention mechanisms at the feature and view level.
    It is composed of an embedding layer that allows the generation of dense and smaller vector representations of the data.
    The feature-level attention mechanism allows the embedding vectors to be weighted for each view.
    The next layer models pairwise interactions between the views by calculating the sum of Hadamard products between the vectors of each view.
    This extends the number $n$ of views by $\frac{n(n - 1)}{2}$ interaction views.
    The attention mechanism at the view level then weights the new generated views to pay more or less attention to each view according to their relevance.
    \item NAML model \cite{wu_neural_2019}:
    This model is called News recommendation Approach with attentive Multi-view Learning.
    It uses a news article encoder with multi-view learning.
    This encoder is used to create an abstract representation of news articles.
    Hierarchical attention mechanisms are used within the encoder for features and views.
    Another attention mechanism is also used to combine representations of the multiple articles the user has viewed.
\end{itemize}

In this paper we will present some of these state-of-the-art models to apply them to our industry data and compare their performance in the context of our recommender system.

\subsection{Attention mechanisms}

The first attention mechanism in deep learning was introduced in 2014 in \cite{bahdanau_neural_2016}.
It was initially developed to improve the performance of encoder-decoder models over long sentences.

Encoder-decoders are composed of two recurrent neural networks, the first one encodes a sequence into a latent vector representation, this vector is called a context vector.
The second uses this abstract representation of the initial sequence to generate a new output sequence.
In such a model, the terms of the input sequence are propagated one after the other through the network.
After several terms have been propagated, the model tends to forget the oldest terms because of the problem of vanishing gradient.
The input sequence is encoded within a context vector, this vector is an embedding of fixed size and the decoder only has access to it to generate the new sequence.
If some of the information has been forgotten or is missing within the context vector, then the quality of the output sequence will not be optimal.

To solve this problem, Bahdanau et al. \cite{bahdanau_neural_2016} proposed an encoder-decoder in which the decoder not only relies on a single context vector to create the new sequence, but rather generates a new context vector containing the information needed at each step.
In this model the decoder has access to all the hidden vectors of the encoder ($h_1, ..., h_n$).
To generate each term of the new sequence the decoder relies on the last generated term $y_{i-1}$, on the previous decoder hidden vector $s_{i-1}$ and on a context vector $c_i$ capturing the relevant information of the input sequence for step $i$.
A formal notation of the generation of the $i$-th term $s_i$ by the decoder is the following recursive function :
\begin{equation}
    s_i = f(s_{i-1}, y_{i-1}, c_i)
\end{equation}
The context vector $c_i$ captures the relevant information for the $i$-th decoding step within the hidden encoder vectors and the previous hidden decoder vector.
A score $e_{i,j}$ is calculated for each hidden encoder vector $h_j$, this score will be used to weight each hidden vector according to its relevance for the $i$-step:
\begin{equation}
    e_{i,j} = a(s_{i-1}, h_j)
\end{equation}
In this notation $a$ is an alignment model, such as neural network layer.
The parameters of the alignment model are trained simultaneously during the optimization of the global model.
The obtained sequence of scalars $e_{i,1}, ..., e_{i,n}$ is normalized with a softmax function whose terms are defined as follows:
\begin{equation}
    \alpha_{i,j} = \frac{exp(e_{i,j})}{\sum_{k=1}^{n}exp(e_{i,k})}
\end{equation}
The obtained vector $\alpha_{i}$ is called the alignment vector and is used to weight the hidden encoder vectors to generate the context vector $c_i$:
\begin{equation}
    c_i = \sum_{j=1}^{n}\alpha_{i,j} \cdot h_j
\end{equation}
This vector contains the relevant information from the input sequence for the $i$-th decoding step.

Attention mechanisms can be applied at different levels in order to extract interesting and relevant information at each of these levels \cite{yang_hierarchical_2016}.
This theory is called hierarchical attention and has been successfully applied to capture basic notions about the structure of a document.
Documents have a hierarchical structure, with words forming sentences that form paragraphs that form a document.
Not all words have the same importance within a sentence and therefore should not be considered equal by the model, the same applies to the importance of sentences and paragraphs.
A model that includes a hierarchical attention mechanism is therefore able to take into account this notion of various importance at different levels.

In this paper we will first describe multi-view learning methods applied to recommender systems which will be evaluated on our industry data in a comparative study.
We will then detail the used data as well as the experimental protocol used for the study.
Finally, we will present the results obtained in our comparative study.

\section{Multi-view learning for recommendation}

\subsection{Available views}

Rezulteo is an online comparison tool that contains a recommender system, this system is subject to constraints, such as the cold-start problem \cite{schein_methods_2002}, implicit interactions between the user and the products, and no purchase confirmation after product recommendation (the user is redirected towards retailer sites).

This section focuses on the detailed presentation of the five available views for the Rezulteo project and on their individual evaluation.
All the models that will be presented will exploit all or a part of those data sources to produce their recommendation.

\begin{enumerate}
    \item \text{User sessions data: }
    This data is obtained from the user's browser session history.
    It is sequential data, since it brings together all actions taken by one user after another, in chronological order, thus its temporal dimension is primordial.
    
    \item \text{Expert product data: }
    This is business data collected from manufacturers that characterizes each Rezulteo product.
    It is static data that associates a unique and abstract representation to each product.
    
    \item \text{Latent user vectors: }
    Latent user vectors are generated by factoring matrices on user interactions using the eALS model, Element-wise Alternating Least Square model \cite{he_fast_2016}.
    The goal is to reuse one of the two components (user matrix) needed to reconstruct the implicit interaction matrix to generate a unique and abstract representation for each user.
    
    \item \text{Comparability view: }
    This static view is obtained from similarity coefficients which are calculated between each pair of products.
    These are determined by an expert system based on business indicators (age of the product, market sales volume and its presence on the different market shares, etc.).
    To make it exploitable, a vector of size $n$ is associated with each product, $n$ being the number of products in total.
    Each product is assigned a unique index between $0$ and $n - 1$, which is determined arbitrarily and represents the position of the product within the list.
    The vector associated with each product contains the similarity values between that product and all other products in the system.
    The values are organized within the vectors according to the indices of the corresponding products in order to make it possible to learn about this data.
    
    \item \text{Compatibility view: }
    This static view lists all the products that are compatible with user sessions, each session being associated with a query that determines this compatibility.
    As each vehicle is compatible with only a limited set of tires, this view simply represents the list of products that are compatible with the user session.
    It is a simple vector of boolean values of size $n$, where $n$ is the total number of products.
\end{enumerate}

The quality of each view has been analyzed by training a neural network on each isolated view.
The architecture of the used neural network is always the same, so that if the obtained recommendation quality is higher than the one obtained using randomly generated data, it means that the isolated view contains useful information for recommendation.

By using several views, all of which provide useful information, a deep composite model is able to naturally make use of the complementary principle.
This principle defines that multiple views that all contain useful and complementary information can be exploited to lead to better learning performance than if they were treated independently of each other.
A deep composite model using multiple views is also capable of making use of the consensus principle.
This principle defines that by minimizing the disagreement between each view, the error rate on each view will also be minimized.
Such a model is capable of using this principle since each view is associated with an independent branch in the model architecture.
Thus, each branch learns from one view, then all the branches are combined into one.
In the final branch, the abstract representations of each view are brought together to retrieve all the useful and complementary knowledge they contain and exploit them together to maximize learning performance.

To evaluate our views we used two common metrics in the recommendation field which complement each other :
\begin{enumerate}
    \item \text{HR (Hit Rank),} it measures the system capacity to provide all relevant solutions (varies from 0 to 100\%).
    \begin{equation}
        HR = \frac{hits}{users}
    \end{equation}
    $hits$ : the number of users for which the product has properly been recommended.\\
    $users$ : the total number of users.
    \item \text{NDCG (Normalized Discounted Cumulative Gain),} it measures the quality of the produced ranking (varies from 0 to 100\%).
    \begin{equation}
    \begin{split}
        DCG_p &= \sum_{i=1}^p \frac{rel_i}{log_2(i + 1)}\\
        IDCG_p &= \sum_{i=1}^{rel_p} \frac{2^{rel_i}}{log_2(i + 1)}\\
        NDCG_p &= \frac{DCG_p}{IDCG_p}
    \end{split}
    \end{equation}
    $p$ : The position in the ranking for which the gain is calculated.\\
    $rel_i$ : The graded relevance of the result at position $i$.
\end{enumerate}

Table \ref{tab:views_evaluation} shows the quality of each available view for this project in comparison to a view composed of random data.
Each value is obtained by the mean of 10 experiments of 20 epochs each.
The metrics are both calculated on a top 100 and all values are converted to percentages.

\begin{table}[ht]
\begin{center}
    \caption{Evaluation of each available view in comparison to a random view.}
    \begin{tabular}{|l|c|c|}
        \hline
        \textbf{Evaluated view} & \textbf{HR@100} & \textbf{NDCG@100}\\
        \hline
        \textbf{Random} & $66.33\pm0.08$ & $19.83\pm0.04$\\
        \hline
        \textbf{User sessions} & $82.01\pm0.30$ & $32.42\pm0.36$\\
        \hline
        \textbf{Expert data} & $83.58\pm0.10$ & $34.16\pm0.13$\\
        \hline
        \textbf{Comparability view} & $83.83\pm0.13$ & $31.82\pm0.13$\\
        \hline
        \textbf{Compatibility view} & $85.37\pm0.17$ & $33.89\pm0.18$\\
        \hline
        \textbf{Latent user vectors} & $89.56\pm0.10$ & $37.21\pm0.18$\\
        \hline
    \end{tabular}
    \label{tab:views_evaluation}
\end{center}
\end{table}

The user sessions data cannot be processed with the same neural network architecture as the other views, as it is sequential data.
The session view is therefore evaluated using a 3D CNN (3D Convolutional Neural Network), a convolution model that is able to handle temporal data using three-dimensional convolutions, it has already been used for this purpose by the authors of \cite{tuan_3d_2017}.
The used confidence interval is the standard deviation.

The evaluation of the quality of the views shows that all available views provide useful information to make recommendation.
As the evaluation is based on a top 100, the random view obtains a HR and NDCG that might seem high, but we can see significantly better results for the other views.
Deep composite models will therefore be able to exploit the consensus principle on the different views to obtain the best possible results.

\subsection{Evaluated models}

This section focuses on the different state-of-the-art models that we have implemented, adapted and applied to our industrial data within the framework of the comparative study.
Our goal is to study multi-view learning performance when applied to the field of industrial data recommendation.
We chose three state-of-the-art models that will be compared to our Baseline, a hybrid between eALS and a 3D CNN.
The model architectures are described and presented in a generic way in order to allow their application on different data.

Some of the models previously presented could not be applied to our industrial data or were too similar to other implemented models, thus they were not selected to be part of our comparative study.

\subsubsection{Baseline}

The used Baseline is based on the work of \cite{burke_hybrid_survey} from which we used the "hybridization by feature combination" to build this model.
Its architecture is composed of a 3D-CNN, a convolutional model capable of handling sequential data.
The input to this model is a concatenation of the user sessions data, expert data and latent user vectors.
Therefore, it only makes use of three out of the five available views.
The views concatenation is carried out so as to keep the temporal dimension of the user sessions data.

\subsubsection{MV-DNN, Multi-View Deep Neural Network \cite{elkahky_multi-view_2015}}

This model is capable to handle data organized in multiple views, it is composed of as many parallel branches as there are used different views.
Each branch of the model handles one view and builds a dense and abstract representation of that view.
The model possesses a pivot view (corresponding to the users data in the paper \cite{elkahky_multi-view_2015}) and aims to maximize the sum of the similarities between the pivot view and all other views by using a cosine similarity function.

\begin{figure}[ht]
\begin{center}
    \includegraphics[width=\linewidth]{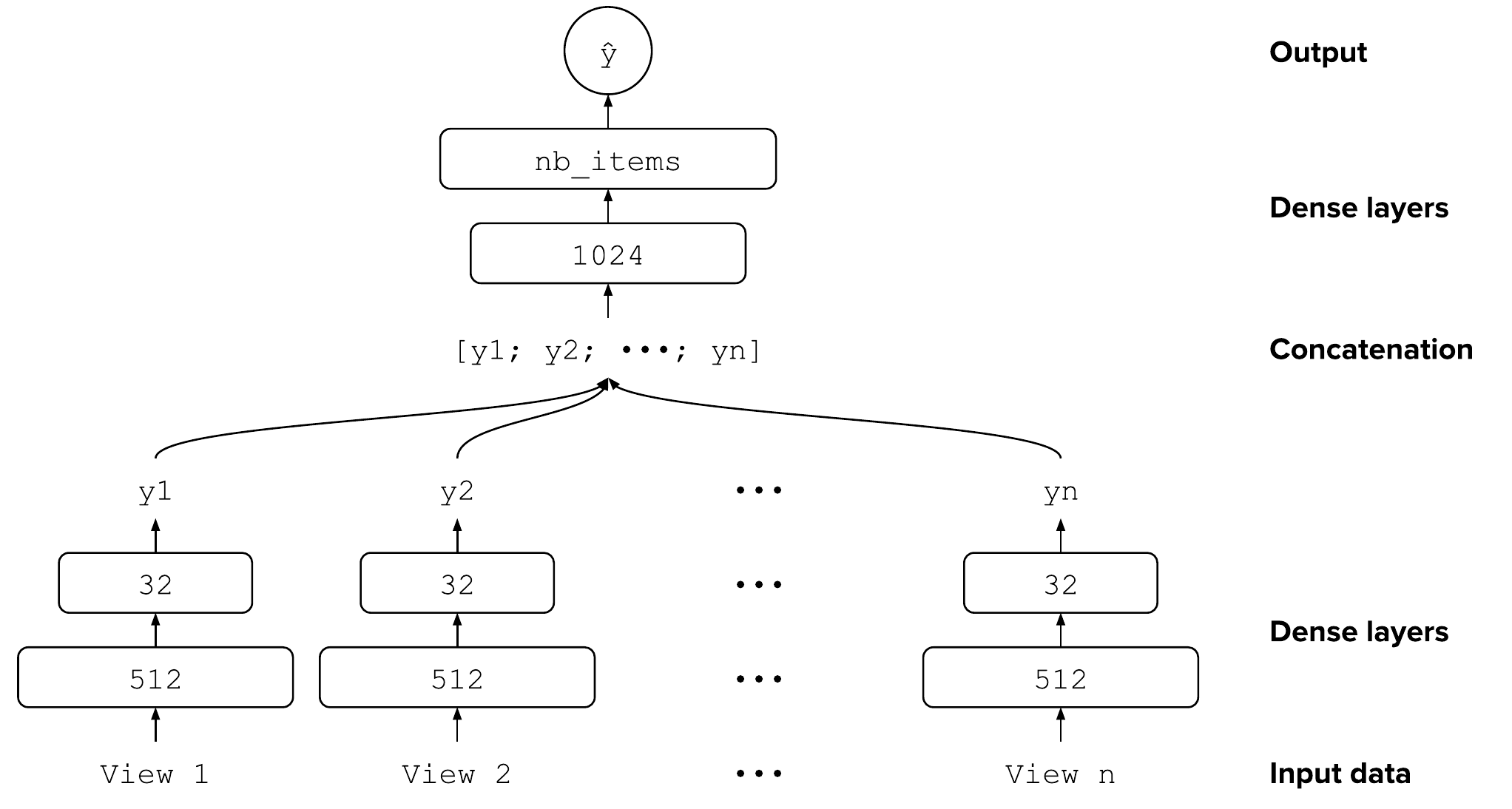}
    \caption{Architecture of our implemented version of the MV-DNN.}
    \label{fig:mv_dnn}
\end{center}
\end{figure}

For the sake of generalization and adaptation to our project, the implemented version of this model, showed on figure \ref{fig:mv_dnn}, does not make use of a pivot view and a similarity function.
In our version the results obtained from each branch are concatenated to be used as input to a final neural network.
The output of the model is a vector associating a recommendation weight to each Rezulteo product.

The developed model architecture is generic and can handle as many views as necessary.
All the branches are made up of forward propagation neural networks with the same architecture, so each branch is able to manage one static view.
The user sessions view is composed of sequential data, therefore it cannot be exploited by this model.

\subsubsection{TDSSM, Temporal Deep Semantic Similarity Model \cite{song_multi-rate_2016}}

The previously presented MV-DNN has the inconvenience of only being able to manage static data.
In order to make use of the total available data for this project it is mandatory for us to use a model capable of handling both static and sequential views.

\begin{figure}[ht]
\begin{center}
    \includegraphics[width=\linewidth]{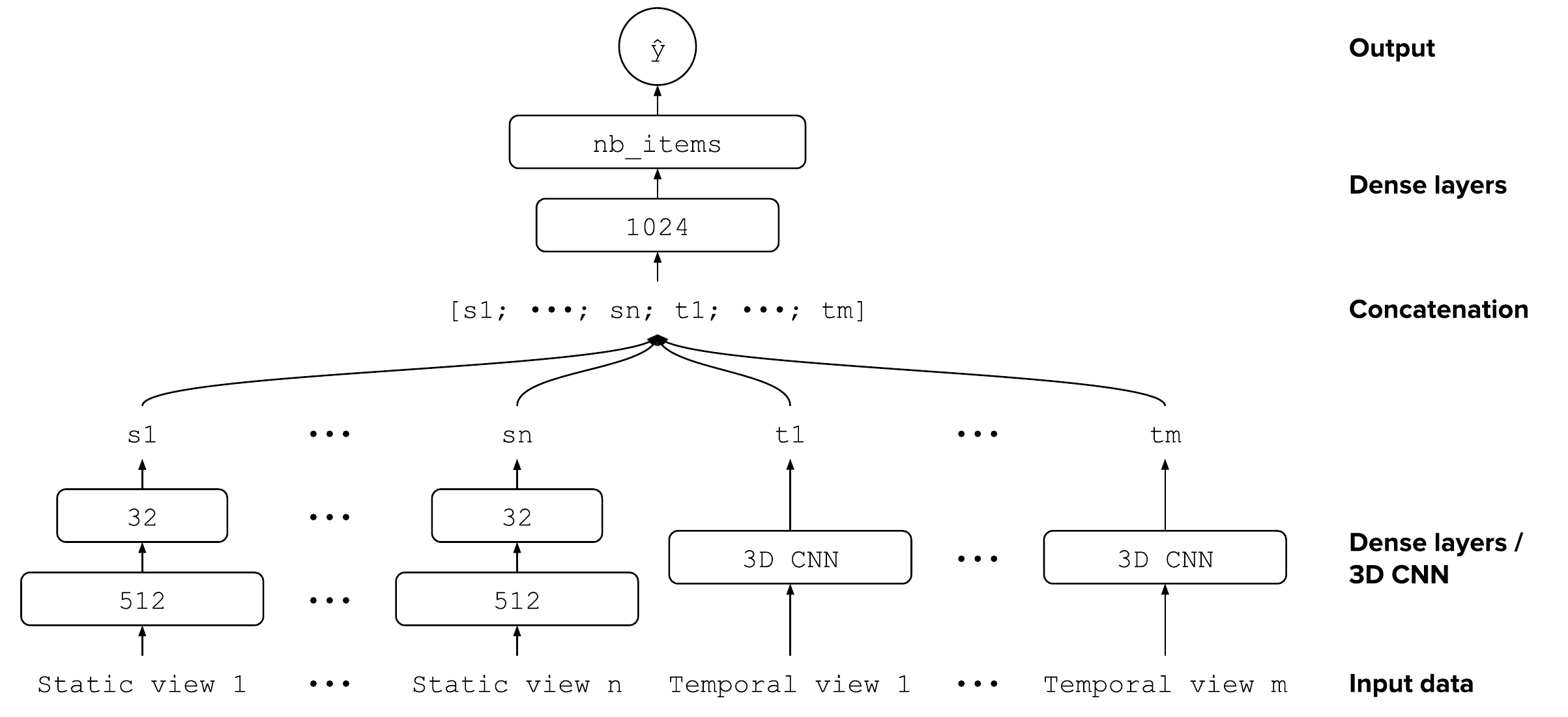}
    \caption{Architecture of our implemented version of the TDSSM.}
    \label{fig:tdssm}
\end{center}
\end{figure}

The TDSSM model extends the MV-DNN architecture by integrating one or more recurrent models within its branches.
Rather than using a recurrent model, our implementation uses a 3D-CNN.
As for the previous model, the implemented architecture is generic so that it can manage static and sequential views, it is represented on figure \ref{fig:tdssm}. 

\subsubsection{MV-AFM, Multi-View Attentional Factorization Machines \cite{liang_multi-view_2020}}

\begin{figure}[ht]
\begin{center}
    \includegraphics[width=\linewidth]{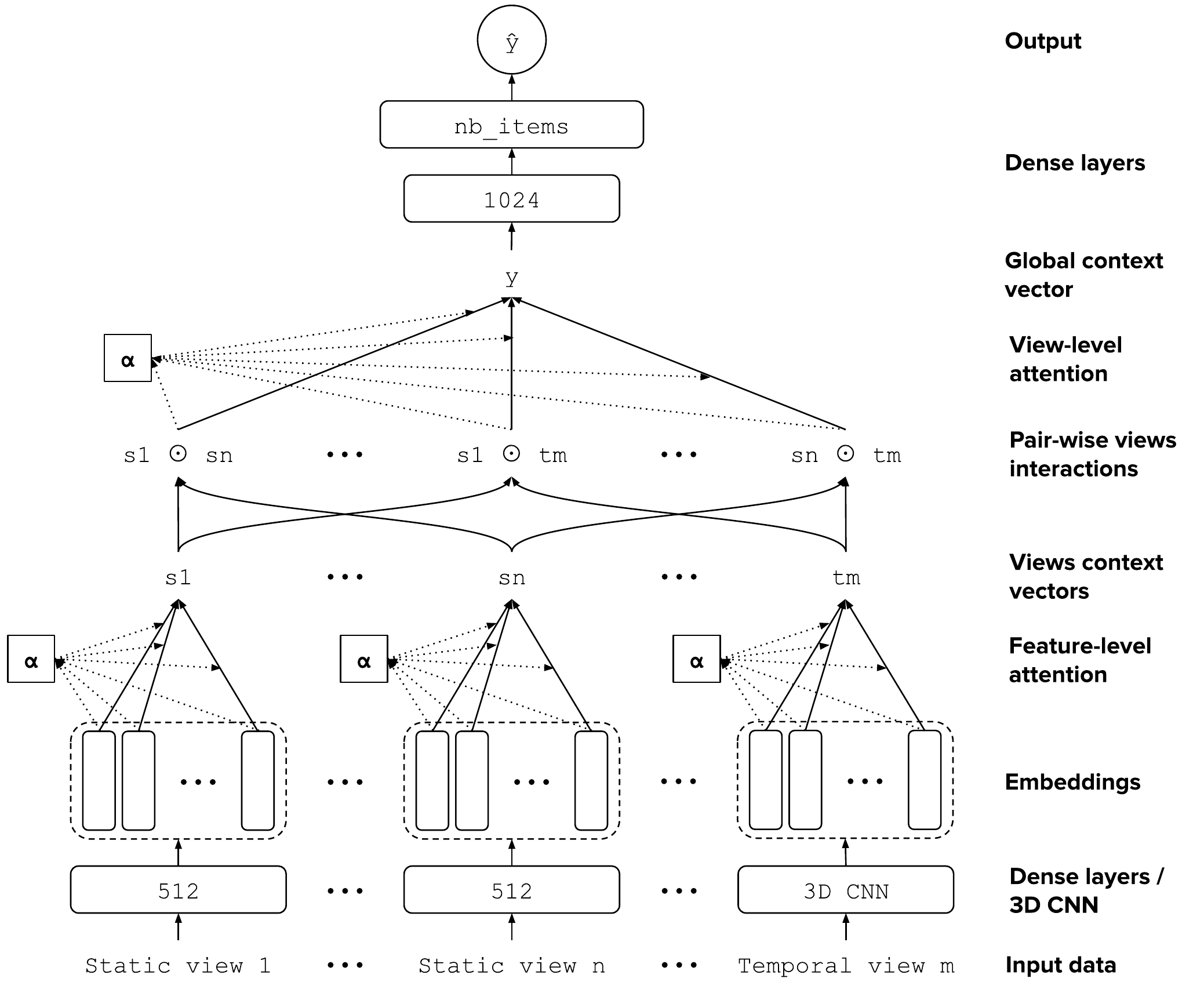}
    \caption{Architecture of our implemented version of the MV-AFM.}
    \label{fig:mv_afm}
\end{center}
\end{figure}

The MV-AFM model introduces two new concepts to multi-view architectures, the hierarchical attention concept and the concept of pairwise interactions between the views.
The purpose of those two additions is to improve the learning capabilities of the model.
Using attention mechanisms is an increment to the work previously done on the Rezulteo project in order to better manage the multi-view context.
The pairwise combination of the views can theoretically allow correlating information between the views and thus improve recommendation performance.
In the initial paper \cite{liang_multi-view_2020}, the model was applied in a multi-view context on datasets from Google Play and from the Apple App Store.
The model architecture has been slightly modified to be fully adapted to the Rezulteo project, it is shown on figure \ref{fig:mv_afm} and described more precisely below.

The model architecture is organized hierarchically, with a feature-level attention mechanism and a global view-level attention mechanism.
There is one branch per view, each branch starts with a model specified to handle the input data type (static or sequential), which returns a fixed size vector.
That vector goes through an embedding layer, which generates $x$ vectors of size $y$ by passing the input vector through $x$ linear layers in parallel, each of $y$ neurons.
Those $x$ vectors go through the feature-level attention mechanism which generates a context vector that theoretically contains the most relevant and useful information from each of the $x$ entry vectors.
The next layer models pairwise view interactions, it operates a Hadamard product between each context vector pair, which makes it possible to model the interactions between all pairs of views one by one.
The view-level attention mechanism can then generate a global context vector that brings together all the relevant information from all the views and all their interactions.
All branches are thus gathered into one and the final output of the model is obtained through two final linear layers.

The feature-level attention mechanism takes the matrix $F$ as input, $F$ is composed of the $x$ vectors of the size $y$ previously generated by the embedding layer.
The purpose of this attention mechanism is to extract the most relevant knowledge from the input in order to generate a context vector from it.
The realized experiments on this model have shown better performances with this attention mechanism than without.
The mechanism can be formalized by the following \ref{eq:attention}.
\begin{equation}
\begin{split}
    e_i &= q \cdot tanh(W \cdot F_i + b)\\
    \alpha &= \frac{exp(e_i)}{\sum_{k=1}^{x} exp(e_k)}\\
    c &= \sum_{i=1}^{x} \alpha_i \cdot F_i
\end{split}
\label{eq:attention}
\end{equation}
With $alpha$ the alignment vector of size $x$ and $c$ the generated context vector of size $y$.
The variables $q$, $W$ and $b$ are weights and biases that are learned during the global training of the model.

The pairwise view interaction layer takes as input the context vectors obtained with the feature-level attention mechanism from each branch, thus there is one context vector per input view.
The point of this layer is to model the interactions between each pair of views.
The resulting number of views is $o + \frac{o\cdot(o - 1)}{2}$, with $o = n + m$, $n$ being the number of static input views and $m$ the number of temporal input views.
Each interaction view is generated by the sum of the Hadamard product between two context vectors.
The size of the output vectors is the same as the size of the input vectors, $y$.
The output of this layer are the $o$ input vectors and the $\frac{o\cdot(o - 1)}{2}$ newly generated vectors, on figure \ref{fig:mv_afm} the $o$ input vectors do not appear in the layer output for the sake of clarity.

The view-level attention mechanism extracts the most useful and pertinent knowledge from all the views and their pair interactions.
It takes as input the $o + \frac{o\cdot(o - 1)}{2}$ vectors from the previous layer and outputs one context vector of size $y$.
This attention mechanism is based on the same model as the feature-level one and can therefore be formalized in the same way \ref{eq:attention}.
The generated context vector is then given as input to two last forward propagation layers that output the final recommendation vector.

\section{Industrial data and experimental protocol}

We conducted our experiments on real data from an online tire comparison tool.
We have two datasets at our disposal.
The first one, noted D1, is composed of four views (user sessions data, latent user vectors, expert data and the comparability view), only the compatibility view is missing.
The second one, noted D2, is a subset of the first one and is composed of the five available views.
The compatibility view, missing from D1, is built for D2 by re-executing user queries to obtain the list of all the products that are compatible with the query.
Table \ref{tab_vol} summarizes the volume of the two datasets:
\begin{table}[ht]
\begin{center}
    \caption{Datasets volume.} \label{tab_vol}
    \begin{tabular}{|l|c|c|c|c|}
        \hline
        Dataset & Sessions & Products & Interactions & Users\\
        \hline
        D1 & $114,359$ & $7,726$ & $307,231$ & $102,613$ \\
        \hline
        D2 & $50,241$ & $3,268$ & $144,095$ & $46,148$ \\
        \hline
    \end{tabular}
\end{center}
\end{table}

From the datasets volume it can be seen that there is about one session per user, which means that users usually only use the recommender system once, and therefore it is impossible for us to create a user profile.
Without user profiles our recommender system is subject to the cold-start issue, which means that the system has to recommend product to a user for whom it has no previous information.
Furthermore, the data that is collected during the system utilization is implicit interactions between the user and the system, thus in our case it is not possible to rely on explicit user ratings or feedback as is usually the case with recommender systems.
It should also be noted that the system does not collect information regarding the purchase of a product by the user, since the user is redirected to the merchant site if he is interested in a product, which further complicates the recommendation.

Our goal is to evaluate the four selected model performances on our two available datasets.
Our Baseline will serve as a benchmark for comparison.
The metrics used to evaluate the models are the same as those used in the previous evaluation of the views: HR and NDCG.
Each interaction between the user and Rezulteo concerns a product, the target variable that the models learn to recommend is the product concerned by the next user interaction.

In order to keep a consistent comparison between the two datasets, we balanced the size of the ranking to be recommended depending on the number of products included in each dataset.
The dataset D1 contains about $2.36$ times as many products as the second, thus models using D1 will be evaluated on a top $236$ and a top $12$, while those using D2 will be evaluated on a top $100$ and a top $5$.

\section{Comparative study}

Table \ref{tab_parameters} shows the number of trainable parameters for each selected model, depending on the used dataset, therefore depending on the number of products that can be recommended.

\begin{table}[ht]
\begin{center}
    \caption{Count of trainable parameters per model and per dataset.}
    \begin{tabular}{|l|c|c|c|c|}
        \hline
        Dataset & Baseline & MV-DNN & TDSSM & MV-AFM \\
        \hline
        D1 & $8,094,087$ & $12,036,431$ & $12,187,687$ & $10,490,119$ \\
        \hline
        D2 & $3,524,637$ & $6,907,973$ & $7,059,229$ & $5,752,477$ \\
        \hline
    \end{tabular}
    \label{tab_parameters}
\end{center}
\end{table}

Tables \ref{tab_D1} and \ref{tab_D2} show the performance of the models evaluated on datasets D1 and D2.
Bold values are the highest values for each metric.
Each value is obtained by the mean of 20 experiments for which the seed of the stochastic gradient descent varies.
Each experiment is performed for 50 epochs for the Baseline and 10 epochs for the other models.
Indeed, multi-view models converge faster than the Baseline, thus the training time of those models is shorter.

\begin{table}[ht]
\begin{center}
    \caption{Models evaluation using dataset D1.}
    \label{tab_D1}
    \resizebox{\columnwidth}{!}{
    \begin{tabular}{|l|c|c|c|c|}
        \hline
        Metric    & Baseline & MV-DNN & TDSSM & MV-AFM \\
        \hline
        HR@236      & $92.12\pm0.13$ & $93.45\pm0.14$ & $93.39\pm0.16$ & $\mathbf{93.47\pm0.21}$ \\
        \hline
        HR@12       & $49.04\pm0.40$ & $51.36\pm1.50$ & $\mathbf{52.02\pm1.02}$ & $51.11\pm0.65$ \\
        \hline
        NDCG@236    & $34.64\pm0.21$ & $\mathbf{35.12\pm0.97}$ & $35.06\pm0.76$ & $34.55\pm0.36$ \\
        \hline
        NDCG@12     & $26.36\pm0.26$ & $26.97\pm1.23$ & $\mathbf{26.98\pm0.96}$ & $26.29\pm0.43$ \\
        \hline
    \end{tabular}
    }
\end{center}
\end{table}

\begin{table}[ht]
\begin{center}
    \caption{Models evaluation using dataset D2.} 
    \label{tab_D2}
    \resizebox{\columnwidth}{!}{
    \begin{tabular}{|l|c|c|c|c|}
        \hline
        Metric & Baseline & MV-DNN & TDSSM & MV-AFM \\
        \hline
        HR@100      & $88.51\pm0.15$ & $91.79\pm0.17$ & $91.82\pm0.12$ & $\mathbf{91.91\pm0.22}$ \\
        \hline
        HR@5        & $33.18\pm0.28$ & $40.13\pm0.26$ & $40.22\pm0.38$ & $\mathbf{40.33\pm0.54}$ \\
        \hline
        NDCG@100    & $35.30\pm0.17$ & $39.84\pm0.16$ & $\mathbf{39.87\pm0.17}$ & $39.57\pm0.38$ \\
        \hline
        NDCG@5      & $21.93\pm0.22$ & $27.09\pm0.22$ & $\mathbf{27.13\pm0.31}$ & $26.78\pm0.49$ \\
        \hline
    \end{tabular}
    }
\end{center}
\end{table}

From these results it can be observed that the three models using the multi-view approach obtain better results than the Baseline.
This shows that the multi-view approach is relevant and effective for our project.
We can notice that the MV-AFM model obtains slightly better results than the others for the Hit Rate metric, and that the TDSSM gets better results for the NDCG metric.
It may therefore be appropriate to use the MV-AFM if one wishes to maximize the chances of recommending the most coherent product and to use the TDSSM if one wishes to achieve a more relevant ranking.
However, given the very narrow range of performance between the three multi-view models it is not possible to say for sure that one of those models is superior to the others.

In order to be able to distinguish the recommendation performance of the three multi-view models and to investigate why the MV-AFM seems to perform better on the Hit Rate and the TDSSM on the NDCG metric, it would be necessary to conduct a more thorough study that would compare their performance on various datasets and not only on a real case as was the case in this study.

\begin{figure}[ht]
\begin{center}
    \includegraphics[width=\linewidth]{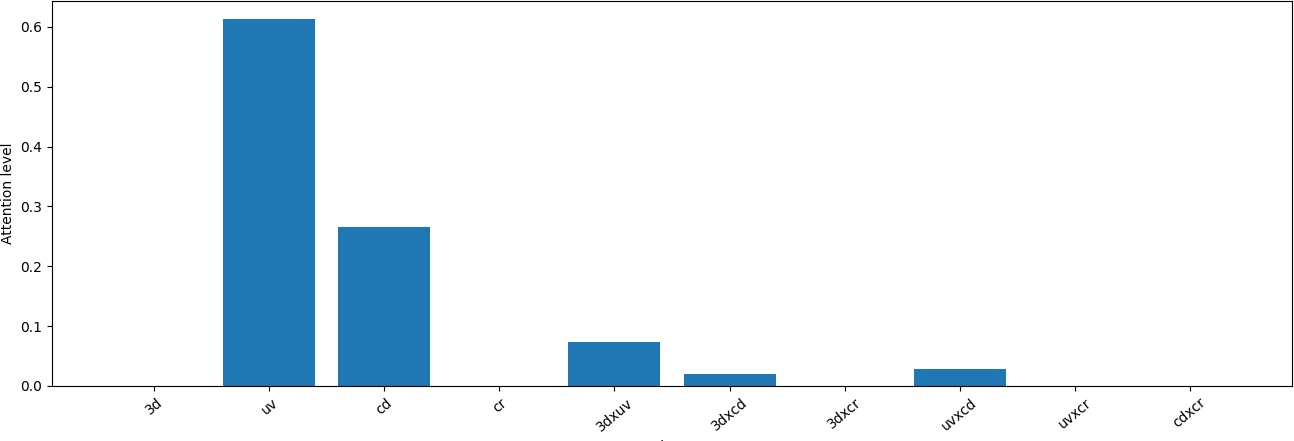}
        \caption{Average attention levels for the MV-AFM on dataset D1.}
    \label{fig:D1_att}
\end{center}
\end{figure}

\begin{figure}[ht]
\begin{center}
    \includegraphics[width=\linewidth]{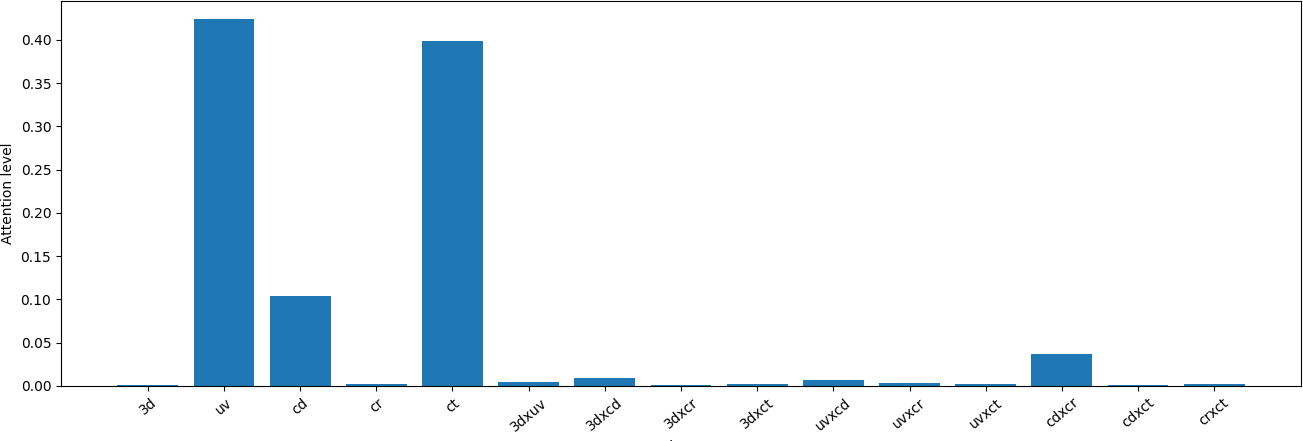}
    \caption{Average attention levels for the MV-AFM on dataset D2.}
    \label{fig:D2_att}
\end{center}
\end{figure}

Figures \ref{fig:D1_att} and \ref{fig:D2_att} show the attention level associated with each view and views interaction within the MV-AFM model, averaged over 1000 predictions.
The y-axis represents the attention level in percentage and the x-axis shows the views and their interactions.
The views are noted as follows:
\begin{itemize}
    \item 3d : User sessions view, handled by the 3D-CNN.
    \item uv : Latent user vectors.
    \item cd : Expert data.
    \item cr : Comparability view.
    \item ct : Compatibility view.
    \item Interactions between views are noted ``view 1 x view 2''.
\end{itemize}

It can be noted that, in both cases, the majority of attention is paid to the latent user vectors (uv), which is in line with the previously conducted evaluation of views that had achieved the best results with this view.
In the case of D1 the expert data (cd) are also widely exploited and the sequential view of user sessions seems to be of interest when correlated with user vectors (3dxuv) and expert data (3dxcd).
In the case of D2 we find a similar pattern with the addition of the compatibility view (ct) which is logically of capital importance in determining recommendations that are consistent with the user's query.

\section{Conclusion}

In this paper, we presented three state-of-the-art models making use of the multi-view learning approach and applied to tire recommendation.
We presented the results of a comparative study between three multi-view models and our Baseline applied to real industrial data.
We observed that the models using the multi-view approach obtained better results than our Baseline, thus demonstrating the relevance of the multi-view approach for recommendation.
The TDSSM model showed the best results for the NDCG metric, which shows its ability to generate a relevant recommendation ranking, while the MV-AFM, using attention mechanisms, obtained the best results for the Hit Rate metric, which shows a higher accuracy in recommending the target product.
Further experiments are planned to evaluate the characteristics of the models that explain a better result on one metric or another.
The compared models managed to achieve a remarkable recommendation quality given the constraints the system faces (cold-start problem, implicit interactions and no purchase confirmation).

\bibliographystyle{IEEEtran.bst}
\bibliography{main}

\begin{thebibliography}{10}
\providecommand{\url}[1]{#1}
\csname url@samestyle\endcsname
\providecommand{\newblock}{\relax}
\providecommand{\bibinfo}[2]{#2}
\providecommand{\BIBentrySTDinterwordspacing}{\spaceskip=0pt\relax}
\providecommand{\BIBentryALTinterwordstretchfactor}{4}
\providecommand{\BIBentryALTinterwordspacing}{\spaceskip=\fontdimen2\font plus
\BIBentryALTinterwordstretchfactor\fontdimen3\font minus
  \fontdimen4\font\relax}
\providecommand{\BIBforeignlanguage}[2]{{%
\expandafter\ifx\csname l@#1\endcsname\relax
\typeout{** WARNING: IEEEtran.bst: No hyphenation pattern has been}%
\typeout{** loaded for the language `#1'. Using the pattern for}%
\typeout{** the default language instead.}%
\else
\language=\csname l@#1\endcsname
\fi
#2}}
\providecommand{\BIBdecl}{\relax}
\BIBdecl

\bibitem{betru_deep_2017}
B.~T. Betru, C.~A. Onana, and B.~Batchakui, ``\BIBforeignlanguage{fr}{Deep
  {Learning} {Methods} on {Recommender} {System}: {A} {Survey} of
  {State}-of-the-art},'' \emph{\BIBforeignlanguage{fr}{International Journal of
  Computer Applications}}, vol. 162, no.~10, 2017.

\bibitem{zhang_deep_2019}
S.~Zhang, L.~Yao, A.~Sun, and Y.~Tay, ``\BIBforeignlanguage{fr}{Deep {Learning}
  based {Recommender} {System}: {A} {Survey} and {New} {Perspectives}},''
  \emph{\BIBforeignlanguage{fr}{ACM Computing Surveys}}, vol.~52, no.~1, pp.
  1--38, Feb. 2019.

\bibitem{adomavicius_toward_2005}
G.~Adomavicius and A.~Tuzhilin, ``\BIBforeignlanguage{fr}{Toward the next
  generation of recommender systems: a survey of the state-of-the-art and
  possible extensions},'' \emph{\BIBforeignlanguage{fr}{IEEE Transactions on
  Knowledge and Data Engineering}}, vol.~17, no.~6, pp. 734--749, Jun. 2005.

\bibitem{schein_methods_2002}
A.~I. Schein, A.~Popescul, L.~H. Ungar, and D.~M. Pennock,
  ``\BIBforeignlanguage{fr}{Methods and metrics for cold-start
  recommendations},'' in \emph{\BIBforeignlanguage{fr}{Proceedings of the 25th
  annual international {ACM} {SIGIR} conference on {Research} and development
  in information retrieval - {SIGIR} '02}}.\hskip 1em plus 0.5em minus
  0.4em\relax Tampere, Finland: ACM Press, 2002, p. 253.

\bibitem{balabanovic_fab_1997}
M.~Balabanović and Y.~Shoham, ``\BIBforeignlanguage{en}{Fab: content-based,
  collaborative recommendation},'' \emph{\BIBforeignlanguage{en}{Communications
  of the ACM}}, vol.~40, no.~3, pp. 66--72, Mar. 1997.

\bibitem{basu_recommendation_1998}
C.~Basu, H.~Hirsh, and W.~W. Cohen, ``Recommendation as {Classification}:
  {Using} {Social} and {Content}-{Based} {Information} in {Recommendation},''
  in \emph{{AAAI}/{IAAI}}, 1998.

\bibitem{tuan_3d_2017}
T.~X. Tuan and T.~M. Phuong, ``\BIBforeignlanguage{fr}{{3D} {Convolutional}
  {Networks} for {Session}-based {Recommendation} with {Content}
  {Features}}.''\hskip 1em plus 0.5em minus 0.4em\relax ACM Press, 2017, pp.
  138--146.

\bibitem{hidasi_parallel_2016}
B.~Hidasi, M.~Quadrana, A.~Karatzoglou, and D.~Tikk,
  ``\BIBforeignlanguage{en}{Parallel {Recurrent} {Neural} {Network}
  {Architectures} for {Feature}-rich {Session}-based {Recommendations}},'' in
  \emph{\BIBforeignlanguage{en}{Proceedings of the 10th {ACM} {Conference} on
  {Recommender} {Systems} - {RecSys} '16}}.\hskip 1em plus 0.5em minus
  0.4em\relax Boston, Massachusetts, USA: ACM Press, 2016, pp. 241--248.

\bibitem{zhao_multi-view_2017}
J.~Zhao, X.~Xie, X.~Xu, and S.~Sun, ``\BIBforeignlanguage{fr}{Multi-view
  learning overview: {Recent} progress and new challenges},''
  \emph{\BIBforeignlanguage{fr}{Information Fusion}}, vol.~38, pp. 43--54, Nov.
  2017.

\bibitem{xu_survey_2013}
C.~Xu, D.~Tao, and C.~Xu, ``\BIBforeignlanguage{fr}{A {Survey} on {Multi}-view
  {Learning}},'' \emph{\BIBforeignlanguage{fr}{arXiv:1304.5634 [cs]}}, Apr.
  2013.

\bibitem{cheng_wide_2016}
H.-T. Cheng, L.~Koc, J.~Harmsen, T.~Shaked, T.~Chandra, H.~Aradhye,
  G.~Anderson, G.~Corrado, W.~Chai, M.~Ispir, R.~Anil, Z.~Haque, L.~Hong,
  V.~Jain, X.~Liu, and H.~Shah, ``Wide \& {Deep} {Learning} for {Recommender}
  {Systems},'' \emph{DLRS 2016: Proceedings of the 1st Workshop on Deep
  Learning for Recommender Systems}, Jun. 2016.

\bibitem{webb_joint_2019}
A.~M. Webb, C.~Reynolds, D.-A. Iliescu, H.~Reeve, M.~Lujan, and G.~Brown,
  ``Joint {Training} of {Neural} {Network} {Ensembles},''
  \emph{arXiv:1902.04422 [cs, stat]}, 2019.

\bibitem{guo_deepfm_2017}
H.~Guo, R.~Tang, Y.~Ye, Z.~Li, and X.~He, ``{DeepFM}: {A}
  {Factorization}-{Machine} based {Neural} {Network} for {CTR} {Prediction},''
  \emph{Proceedings of the Twenty-Sixth International Joint Conference on
  Artificial Intelligence (IJCAI-17), arXiv:1703.04247 [cs]}, Mar. 2017.

\bibitem{he_neural_2017}
X.~He, L.~Liao, H.~Zhang, L.~Nie, X.~Hu, and T.-S. Chua, ``Neural
  {Collaborative} {Filtering},'' \emph{WWW '17: Proceedings of the 26th
  International Conference on World Wide Web, arXiv:1708.05031 [cs]}, Aug.
  2017.

\bibitem{elkahky_multi-view_2015}
A.~M. Elkahky, Y.~Song, and X.~He, ``\BIBforeignlanguage{fr}{A {Multi}-{View}
  {Deep} {Learning} {Approach} for {Cross} {Domain} {User} {Modeling} in
  {Recommendation} {Systems}},'' in \emph{\BIBforeignlanguage{fr}{Proceedings
  of the 24th {International} {Conference} on {World} {Wide} {Web} - {WWW}
  '15}}.\hskip 1em plus 0.5em minus 0.4em\relax Florence, Italy: ACM Press,
  2015, pp. 278--288.

\bibitem{gao_modeling_2014}
J.~Gao, P.~Pantel, M.~Gamon, X.~He, and L.~Deng,
  ``\BIBforeignlanguage{en}{Modeling {Interestingness} with {Deep} {Neural}
  {Networks}},'' in \emph{\BIBforeignlanguage{en}{Proceedings of the 2014
  {Conference} on {Empirical} {Methods} in {Natural} {Language} {Processing}
  ({EMNLP})}}.\hskip 1em plus 0.5em minus 0.4em\relax Doha, Qatar: Association
  for Computational Linguistics, 2014, pp. 2--13.

\bibitem{song_multi-rate_2016}
Y.~Song, A.~M. Elkahky, and X.~He, ``\BIBforeignlanguage{fr}{Multi-{Rate}
  {Deep} {Learning} for {Temporal} {Recommendation}}.''\hskip 1em plus 0.5em
  minus 0.4em\relax SIGIR '16: Proceedings of the 39th International ACM SIGIR
  conference on Research and Development in Information Retrieval, 2016, pp.
  909--912.

\bibitem{liang_multi-view_2020}
T.~Liang, L.~Zheng, L.~Chen, Y.~Wan, P.~S. Yu, and J.~Wu,
  ``\BIBforeignlanguage{fr}{Multi-view factorization machines for mobile app
  recommendation based on hierarchical attention},''
  \emph{\BIBforeignlanguage{fr}{Knowledge-Based Systems}}, vol. 187, p. 104821,
  Jan. 2020.

\bibitem{wu_neural_2019}
C.~Wu, F.~Wu, M.~An, J.~Huang, Y.~Huang, and X.~Xie, ``Neural {News}
  {Recommendation} with {Attentive} {Multi}-{View} {Learning},''
  \emph{Proceedings of the Twenty-Eighth International Joint Conference on
  Artificial Intelligence (IJCAI-19), arXiv:1907.05576 [cs]}, Jul. 2019.

\bibitem{bahdanau_neural_2016}
D.~Bahdanau, K.~Cho, and Y.~Bengio, ``Neural {Machine} {Translation} by
  {Jointly} {Learning} to {Align} and {Translate},'' \emph{International
  Conference on Learning Representations 2014, arXiv:1409.0473 [cs, stat]}, May
  2016.

\bibitem{yang_hierarchical_2016}
Z.~Yang, D.~Yang, C.~Dyer, X.~He, A.~Smola, and E.~Hovy,
  ``\BIBforeignlanguage{fr}{Hierarchical {Attention} {Networks} for {Document}
  {Classification}},'' in \emph{\BIBforeignlanguage{fr}{Proceedings of the 2016
  {Conference} of the {North} {American} {Chapter} of the {Association} for
  {Computational} {Linguistics}: {Human} {Language} {Technologies}}}.\hskip 1em
  plus 0.5em minus 0.4em\relax San Diego, California: Association for
  Computational Linguistics, Jun. 2016, pp. 1480--1489.

\bibitem{he_fast_2016}
X.~He, H.~Zhang, M.-Y. Kan, and T.-S. Chua, ``\BIBforeignlanguage{fr}{Fast
  {Matrix} {Factorization} for {Online} {Recommendation} with {Implicit}
  {Feedback}}.''\hskip 1em plus 0.5em minus 0.4em\relax ACM Press, 2016, pp.
  549--558.

\bibitem{burke_hybrid_survey}
R.~Burke, ``Hybrid recommender systems: Survey and experiments,'' \emph{User
  Modeling and User-Adapted Interaction}, vol.~12, no.~4, p. 331–370, Nov.
  2002.

\end{thebibliography}

\end{sloppypar}
\end{document}